\documentclass[10pt,twocolumn,letterpaper]{article}
\usepackage{cvpr}
\usepackage{times}
\usepackage{epsfig}
\usepackage{graphicx}
\usepackage{amsmath}
\usepackage{amsthm}
\usepackage{subcaption}    

\def\A{\mathcal{A}}
\def\M{\mathcal{M}}
\def\B{\mathcal{B}}
\def\C{\mathcal{C}}

\def\D{\mathcal{D}}

\def\W{\mathcal{W}}

\def\RR{\mathbb{R}}

\def\x{\mathbf{x}}
\def\y{\mathbf{y}}

\def\fR{\mathfrak{R}}
\def\proj{\mathrm{Proj}_{\mathfrak{R}}}

\newtheorem{theorem}{{\bf Theorem}}

\newtheorem{definition}{{\bf Definition}}

\graphicspath{{figures/}}  

\usepackage{booktabs} 
\usepackage{graphicx}
\usepackage{amsmath}
\usepackage{amssymb}
\usepackage{appendix}
\usepackage{algorithmic}

\usepackage{url}
\usepackage[ruled]{algorithm2e} 

\SetAlFnt{\small}
\SetAlCapFnt{\small}
\SetAlCapNameFnt{\small}
\SetAlCapHSkip{0pt}

\usepackage[pagebackref=true,breaklinks=true,letterpaper=true,colorlinks,bookmarks=false]{hyperref}

 \cvprfinalcopy 

\def\cvprPaperID{7873} 

\begin{document}

\title{A Dual Iterative Refinement Method for Non-rigid Shape Matching}

\author{Rui Xiang\\ 
Department of Mathematics\\
UC Irvine\\
{\tt\small xiangr1@uci.edu}
\and
Rongjie Lai\\
Department of Mathematics\\
Rensselaer Polytechnic Institute\\
{\tt\small lair@rpi.edu}
\and
Hongkai Zhao\\
Department of Mathematics\\
Duke University\\
{\tt\small zhao@math.duke.edu}
}
\maketitle

\begin{abstract}
In this work, a robust and efficient dual iterative refinement (DIR) method is proposed for dense correspondence between two nearly isometric shapes. The key idea is to use dual information, such as spatial and spectral, or local and global features, in a complementary and effective way, and extract more accurate information from current iteration to use for the next iteration. In each DIR iteration, starting from current correspondence, a zoom-in process at each point is used to select well matched anchor pairs by a local mapping distortion criterion. These selected anchor pairs are then used to align spectral features (or other appropriate global features) whose dimension adaptively matches the capacity of the selected anchor pairs. Thanks to the effective combination of complementary information in a data-adaptive way, DIR is not only efficient but also robust to render accurate results within a few iterations. By choosing appropriate dual features, DIR has the flexibility to handle patch and partial matching as well. Extensive experiments on various data sets demonstrate the superiority of DIR over other state-of-the-art methods in terms of both accuracy and efficiency. 
\end{abstract}

\vspace{-0.2cm}
\section{Introduction}
Nonrigid shape matching is one of the most basic and important tasks in computer vision and shape analysis, e.g., shape registration, comparison, recognition, and retrieval. Different from 2D images, shapes are usually represented as 2-dimensional manifolds embedded in $\mathbb{R}^3$. Many challenges for nonrigid shape matching come from embedding ambiguities, i.e. shapes sharing the same metric (isometry) or very similar metrics (nearly isometry) can have drastically different xyz-coordinates representations in $\mathbb{R}^3$. 

To overcome these representation ambiguities, one of the commonly used strategies is to extract features which are isometrically invariant and robust to small perturbations. Along with the idea, many successful and popular approaches are based on spectral geometry ~\cite{Reuter:06,Levy:2006IEEECSMA,Vallet:2008CGF,Bronstein:2010CVPR,lai2010metric,rustamov2007laplace}. Theoretically, the Laplace-Beltrami (LB) operator is isometrically invariant. As a generalization of Fourier basis functions from Euclidean domains to manifolds, the eigensystem of LB operator provides complete intrinsic spectral information of the underlying manifold. 
Moreover, from lower eigen-modes to higher eigen-modes, LB eigen-functions also provide a multi-scale characterization of the underlying manifold from coarse to fine resolution. Although using spectral geometry removed possible non-rigid embedding ambiguities, 
new ambiguities emerge in the spectral domain due to non-uniqueness of the LB eigen-system, e.g., sign ambiguity for eigen-functions, the ambiguity of choosing a basis for the LB eigen-space corresponding to a non-simple LB eigen-value (due to symmetry), the ambiguity of ordering for close eigen-values (due to small perturbations). To handle these ambiguities and use spectral features accurately and robustly, a proper linear transformation (a rigid transformation for exact isometry) 
needs to be found to align the spectral modes between two shapes first. This linear transformation is typically computed through some matching/correlation based on given (prior) correspondence, e.g., landmarks \cite{ovsjanikov2012functional,aflalo2016spectral,lai2017multiscale}. 
Moreover, to resolve fine details and acquire accurate correspondence between two shapes, high eigen-modes need to be used. 
However, the use of higher eign-modes will not only require more computation costs, but, more importantly, can also cause instability with respect to small perturbations.

To tackle the instability issue of using high eigen-modes directly, one natural multi-scale approach is to start from a correspondence at a coarse scale using a few low modes and iteratively refine the correspondence at a finer and finer scale by adding more and more higher modes gradually.
The main motivation is that the linear transformation (a small matrix) on a coarse scale between two truncated spectral spaces spanned by a few low eigen-modes can be determined efficiently and stably from an initial approximate or limited correspondence. Once low eigen-modes are aligned well, an improved correspondence, especially between smooth parts of the two shapes, is likely obtained. The improved correspondence is then used to determine the linear transformation for the next iteration which involves more and higher LB eigen-modes. Such a multi-scale idea for shape correspondence has been proposed in \cite{lai2017multiscale} for multi-scale registration using rotation-invariant sliced-Wasserstein distance and in \cite{melzi2019zoomout} as a Zoom-out process. However, for the above straightforward multi-scale approach in the spectral domain, there are two key issues. First, in each iteration, the determination of the linear transformation between the truncated spectral embedding of two shapes using current correspondence of all points, many of which are incorrect, indiscriminately might be problematic. In Theorem \ref{thm:Error}, we show that the linear transformation between two spectral spaces determined using all points from an inaccurate correspondence will most likely lead to errors. These errors could be very significant and can cause either a failure for later refinement or slow convergence shown in the  supplementary material. The other issue is the lack of a systematic and data-adaptive way to determine how many eigen-modes can be aligned accurately and stably by the current correspondence. Thus, it is hard to decide the appropriate jump in the number of eigen-modes for the next refinement after each iteration to achieve fast convergence. Previously, an increment of one mode was typically used in Zoom-Out \cite{melzi2019zoomout}, while a prefixed sequence of eigenmodes was proposed in \cite{lai2017multiscale} based on the rule of thumb.

\begin{figure}[t]
\centering\includegraphics[width=.9\linewidth]{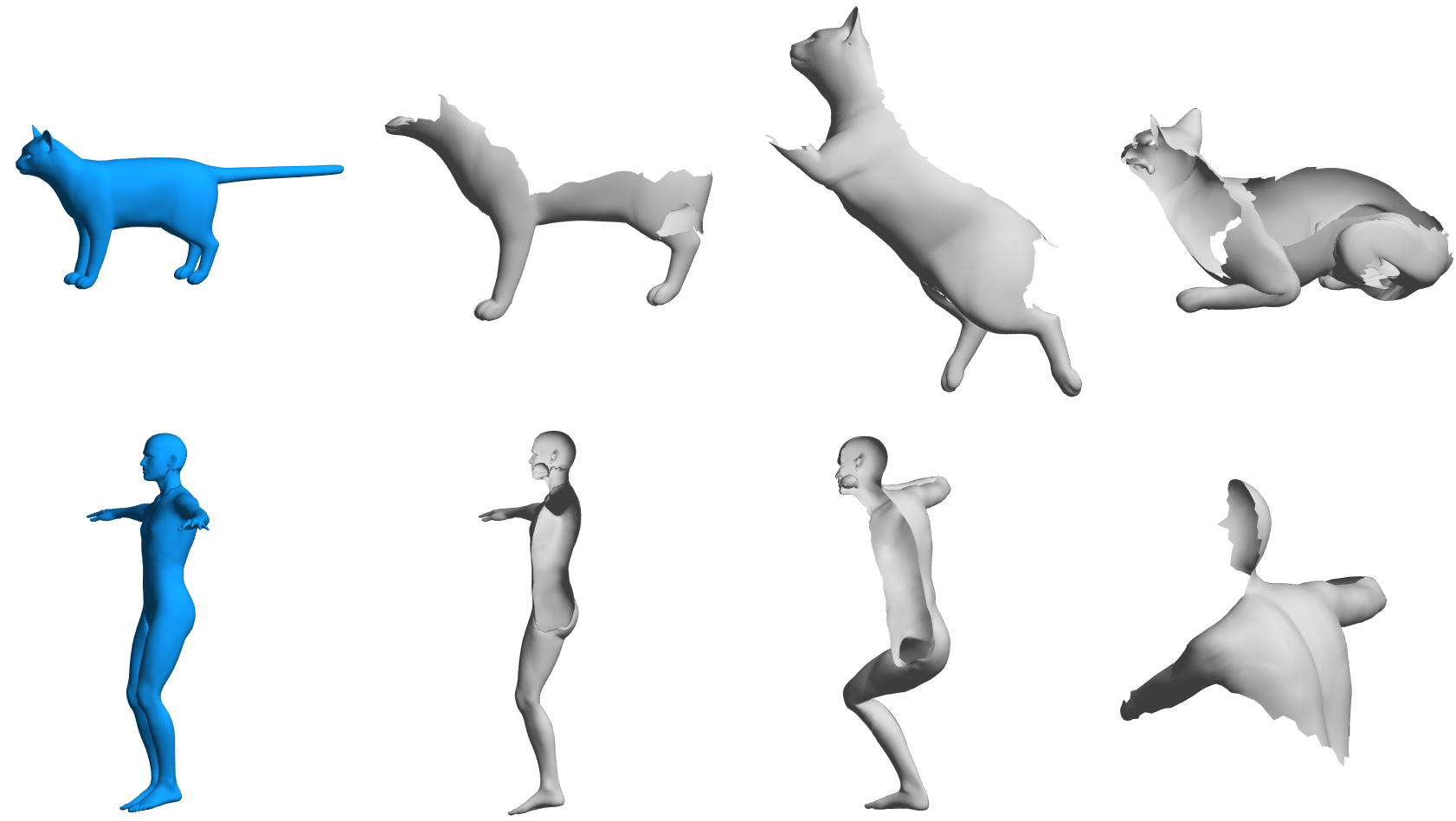}\\
\centering\includegraphics[width=.49\linewidth]{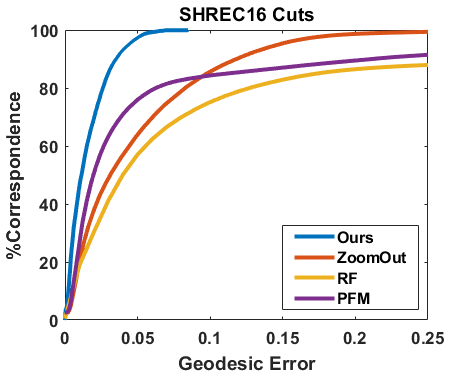}
\centering\includegraphics[width=.49\linewidth]{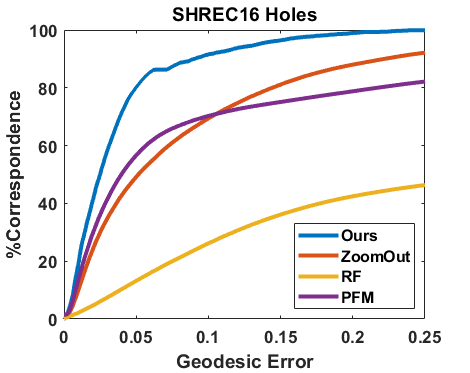}
\caption{\textit{Top:} First row shows examples from SHREC'16 holes, and second row shows examples from SHREC'16 cuts. We map the partial shapes in gray to full shapes in blue. \textit{Bottom:} Geodesic error on SHREC'16 cuts and holes data set with comparison to other state-of-the-art methods.}
\label{SHREC'16}
\vspace{-0.3cm}
\end{figure}


In this work, we propose a simple and efficient dual iterative refinement strategy. This method combines complementary information, such as spatial and spectral, or local and global, to simultaneously refine the correspondence between approximately isometric shapes in an effective way. More specifically, spatial matching via local mapping distortion (see Definition \ref{def:faithfulness}) is applied point-wisely to choose well-matched correspondence -- anchor pairs, at the current stage. Once anchor pairs are selected, they are used to find 1) the maximal dimension of spectral space that can be robustly and accurately determined by these anchor pairs based on their distribution, and 2) the linear transformation that aligns spectral basis at next scale which will lead to a more refined correspondence at next stage. This remarkable simple strategy addresses the aforementioned two critical issues in previous multi-scale approaches using only spectral matching and allows one to utilize all well-matched pairs from the current step to jump to the next step in an accurate, efficient, stable, and data-adaptive way. We use extensive numerical experiments to show that our simple strategy, DIR, outperforms start-of-the-art model-based methods in terms of both accuracy and efficiency markedly. By choosing appropriate and application-specific dual features, DIR enjoys the flexibility to handle different scenarios, such as raw point clouds, patch and partial matching (as illustrated in Figure \ref{SHREC'16}). We defer a detailed explanation of this Figure in Section \ref{sec:Experiments}.


\noindent\textbf{Contributions.} We summarize our main contributions as follows:

\begin{itemize}
\item 
By combining spatial matching via local mapping distortion and spectral matching via LB eigen-basis, the proposed DIR goes from coarse to fine resolution stably and effectively in a data-adaptive way which overcomes the main challenges in spectral matching. 
\item By generalizing this dual refinement strategy to other local and global features such as 
local mapping distortion and global geodesic distance, DIR handles patch and partial matching. 
\end{itemize}

The rest of this paper is organized as follows. In Section \ref{sec:related}, we give a brief review of some related work. Then we present our proposed method, DIR, in detail followed by a discussion about a few possible extensions of our method in Section \ref{sec:DIR}. After that, we conduct extensive experiments on various benchmark data sets and compare DIR to other state-of-the-art methods in terms of both accuracy and efficiency in Section \ref{sec:Experiments}. We conclude the paper in Section \ref{sec:conclusion}.

\section{Related work}
\label{sec:related}
Deigning effective shape descriptors is crucial for shape registration or correspondence. In general, descriptors can be categorized as pointwise or pairwise. Extrinsic pointwise descriptors \cite{tombari2010unique, gumhold2001feature, dubrovina2011approximately, rodola2012game} are easy to compute but usually not very accurate, especially when non-rigid transformation is involved. To deal with non-rigid transformation, different intrinsic descriptors either in spatial domain, such as geodesics distance signatures \cite{van2011survey}, heat kernel signatures \cite{sun2009concise} and wave kernel signatures \cite{aubry2011wave}, or in spectral domain using eigensystem of LB operator~\cite{ Levy:2006IEEECSMA,Vallet:2008CGF,Bronstein:2010CVPR,lai2010metric,ovsjanikov2012functional,lai2017multiscale}, are proposed. Then various nearest neighbor searching or linear assignment methods are used in the descriptor space to find the dense point correspondence. For point-wise descriptors in spectral domain, such as functional map \cite{ovsjanikov2012functional}, a proper linear transformation is needed to align spectral basis to remove ambiguities in the eigensystem of LB operator. The determination of the linear transformation itself is typically based on some prior knowledge, e.g., given landmarks, region correspondence or orientation preserving properties~\cite{ren2018continuous}, and can be challenging to achieve good accuracy especially when the dimension of spectral embedding is high, which is needed to resolve fine shape features to achieve accurate correspondence. On the other hand, spatial domain pointwise descriptors are usually defined smoothly on shapes which is difficult to provide accurate correspondence as well. Our method aims at tackling these issues raised in aligning shapes using pointwise descriptors. 

Intrinsic pairwise descriptors, such as pairwise geodesic distance \cite{bronstein2006, bronstein2010gromov, vestner2017product} and kernel functions \cite{liu2008kernel, shtern2014iterative, vestner2017efficient} have also been proposed for non-rigid shape correspondence problems. Although the matching of pairwise descriptors is stricter and may be more robust and accurate, it requires to solve a quadratic assignment problem (QAP) which is NP-hard. Various methods have been proposed to solve the QAP approximately in a more computational tractable way e.g sub-sampling \cite{tevs2011intrinsic}, coarse-to-fine \cite{wang2011discrete}, geodesic distance sparsity enforcement methods \cite{gasparetto2017spatial} and various relaxation approaches \cite{aflalo2015convex, bronstein2006generalized, kezurer2015tight, leordeanu2005spectral, chen2015robust,dym2017ds++}. 
One popular approach is to relax the nonconvex permutation matrix (representing pointwise correspondence) constraint in the QAP to a doubly stochastic matrix (convex) constraint \cite{aflalo2015convex,chen2015robust}. However, both the pairwise descriptors and the doubly stochastic matrix are dense matrices, which make the relaxed QAP still challenging to solve even for a modest size problem. 
Recently, a novel local pairwise descriptor and a simple, effective iterative method to solve the resulting quadratic assignment through sparsity control was proposed in \cite{xiang2020efficient} for shape correspondence between two approximately isometric surfaces. 



\section{Dual Iterative Refinement Method}
\label{sec:DIR}

\subsection{Functional map}
\label{subsec:FM}
Spectral geometry is widely used in shape analysis \cite{Reuter:06,Levy:2006IEEECSMA,Vallet:2008CGF, Bronstein:2010CVPR,lai2010metric,ovsjanikov2012functional,lai2017multiscale,schonsheck2018nonisometric,han2019orientation}. Leveraging by spectral presentation, functional map is introduced in \cite{ovsjanikov2012functional} to solve non-rigid shape correspondence problem. It improves earlier point-based spectral methods \cite{jain2006, jain2007, mateus2008articulated, ovsjanikov2010} which directly matches the spectral embeddings of shapes. We provide a brief introduction to the concept of functional map which is used in our multi-scale DIR process.

Given $(\M_1,g_1)$ as a closed 2-dimensional Riemannian manifold, the LB operator uniquely determines the underlying manifold up to isometry \cite{berard1994embedding}. 
Eigen-functions of LB operator form an orthonormal basis on the underlying manifold and can be used as intrinsic and multi-scale descriptors for shapes. 
In discrete setting, the manifold $\M_1$ is usually represented as a triangulated mesh with vertices $ \{\x_i\}_{i=1}^n$. 
The LB matrix is given by $\mathcal{L}_{\M_1} = \A_1^{-1} \W_1$ \cite{pinkall93}, where $\A_1$ is the diagonal element area matrix of $\M_1$ and $\W_1$ is the standard cotangent weight matrix. The discrete truncated $k$-dimensional spectral embedding of $\M_1$ can be expressed as a matrix $\Phi_{\M_1}^k\in\mathbb{R}^{n\times k}$ whose rows are the first $k$ LB eigen-functions evaluated at each point. Similarly, we can define the first $k$ LB eigen-embedding for $\M_2$ as $\Phi_{\M_2}^k$. 

The key idea of functional map is to lift a shape correspondence $T:\M_1\rightarrow\M_2$ to a linear map, functional map,  between the function spaces $\C(\M_1)$ and $\C(\M_2)$. The functional map for a given correspondence $T$ can be represented as a linear transformation between two given bases of $\C(\M_1)$ and $\C(\M_2)$, which becomes a matrix when the shapes are discretized and bases are truncated. In practice, LB eigen-functions are commonly used as the basis for the function space. For example, given a permutation matrix $\Pi$ representing a point-to-point map $T$ from $\M_1$ to $\M_2$, the functional map $C$ is defined as:
\begin{equation}
\label{eqn:funmap}
    C = \arg\min_{C} \|\Phi_{\M_1} C  - \Pi \Phi_{\M_2}\|_F^2 
\end{equation}
Theoretically, if $\M_1$ and $\M_2$ are isometric, then the corresponding LB eigen-functions are the same up to possible ambiguities caused by sign switch and non-simple eigen-values, which form an orthonormal group. This motivates a constrained version of the following functional map~\cite{lai2017multiscale}
\begin{equation}
\label{eqn:funmap_ON}
    \min_{C \in \fR} \|\Phi_{\M_1} C  - \Pi \Phi_{\M_2}\|_F^2 =  \proj ( \Phi_{\M_1}^\top \Pi \Phi_{\M_2})
\end{equation}
where $\fR = \{C~|~ C^\top C = Id\}$ denotes the set of orthonormal matrices  and the projection to $\fR$ is provided as $\proj(A) = UV^\top $ with the singular value decomposition (SVD) of $A = U\Sigma V^\top$. 
Since this paper aims at tackling nearly isometric shape matching problem, we adopt \eqref{eqn:funmap_ON} which is equivalent to \eqref{eqn:funmap} for the isometric case. 

Typically, the functional map is obtained via a least square optimization with various constraints and regularizations, such as preservation of given landmarks, communitivity with LB operator, and sparsity~\cite{shtern2014iterative,ren2018continuous}. Once the optimized functional map $C$ is computed, for any $p \in \M$, the correspondence  $T(p)$ can be obtained by solving the following spectral matching problem:
\begin{equation}
\label{eqn:knnmap}
T(p) = \arg\min_{q \in \M_2}\| \Phi_{\M_1}(p) C - \Phi_{\M_2}(q) \|_F^2 
\end{equation}

However, determination of an accurate functional map $C$ can be difficult when limited prior knowledge or a poor correspondence is used. On the one hand, it becomes harder when more eigen-modes are involved since the degrees of freedom of the functional map grow quadratically in terms of the eigen-modes involved and, moreover, high eigen-modes are less computationally stable with respect to perturbations or noises.
On the other hand, confining to low eigen-modes limits resolution of the spectral representation and hence the accuracy of shape correspondence. A natural multi-scale idea is to start the functional map from a coarse resolution involving a few low eigen-modes and construct a coarse correspondence $T$. Then the coarse correspondence is used to help determining the functional map at a finer scale and hence a more refined correspondence. This process can be iterated until a fine enough resolution is achieved. The main motivation is that the functional map at a coarse scale -- a small matrix, can be determined efficiently and stably from an initial approximate or limited correspondence, which is then used to  improve the correspondence at next iteration. This strategy leads to iteratively computing \eqref{eqn:funmap_ON}  and \eqref{eqn:knnmap}, or \eqref{eqn:funmap}  and \eqref{eqn:knnmap}, which is the key idea proposed in \cite{lai2017multiscale} using rotation-invariant sliced Wasserstein distance and in the Zoom-out process \cite{melzi2019zoomout}. 
As a crucial component of these iterative refinement methods, the new functional map $C$ is obtained from the current correspondence of all points. Since intermediate correspondence, especially at the beginning, can be quite inaccurate, this naive strategy may lead to significant errors.

Next, we provide both theoretical and experimental evidences on the effect of correspondence error of $C$ in \eqref{eqn:funmap_ON}. This can be problematic for any iterative refinement procedure based on spectral geometry. Assume we have two perfectly isometric manifold $\M_1, \M_2$ and their corresponding discrete truncated spectral embedding $\Phi_{\M_1},\ \Phi_{\M_2}$. Without loss of generality, we assume $\mathrm{Id}$ is the ground truth correspondence between $\M_1$ and $\M_2$ (otherwise, we can shuffle row vectors of $\Phi_{\M_2}$ according to the ground truth correspondence). The ground truth  functional map is an orthonormal matrix $C_T\in\fR$, i.e.  $\Phi_{\M_2} = \Phi_{\M_1} C_T$ and $C_T = \proj(\Phi_{\M_1}^\top \Phi_{\M_2})$. 
\begin{theorem}
\label{thm:Error}
Given $\Phi_{\M_2} = \Phi_{\M_1} C_T$ with $C_T\in\fR$. Let's assume a one-to-one correspondence $\Pi$ is an inaccurate correspondence which maps a portion of $\Phi_{\M_1}$ accurately to the corresponding  part of $\Phi_{\M_2}$, while the rest part of $\Pi$ is inaccurate. Without loss of generality, 
we write $\Phi_{\M_1} = \begin{pmatrix}
  X_1\\ 
  X_2
\end{pmatrix}$ 
and $\Phi_{\M_2} = 
 \begin{pmatrix}
  Y_1\\ 
  Y_2
\end{pmatrix}$ 
where $X_1,\ Y_1 \in \RR^{n_1 \times k},\ X_2,\ Y_2 \in \RR^{n_2 \times k}$ and $n_1+n_2=n$. We let $\Pi\Phi_{\M_2} = 
 \begin{pmatrix}
  Y_1\\ 
  \sigma Y_2
\end{pmatrix}$ for a permutation matrix $\sigma \in \RR^{n_2\times n_2}$. Let 
$C_a = \arg\min_{C \in \fR} \|\Phi_{\M_1} C  - \Pi \Phi_{\M_2}\|_F^2$.
Then it is most likely that the spectral norm $\|C_a - C_T\|_2 > 0$ with probability at least $1 - \eta$ with $\displaystyle \eta = \sum_{j=0}^{\lfloor n_2/2 \rfloor}\frac{1}{2^j j!(n_2-2j)!}$.
\label{th1}
\end{theorem}
Notice that $\eta$ decreases rapidly as $n_2$ grows. For example, when $n_2 = 25$, $\eta \approx 10^{-12}$. We provide a detailed proof of this theorem in the supplementary material  where we also numerically verify that using points from an inaccurate correspondence may lead to significant errors in the functional map. This can cause failure for iterative refinement.

Another critical issue for multi-scale approach in spectral domain is how to iteratively  increase the resolution, i.e., the dimension of spectral embedding, in an accurate, stable and data adaptive way. Most approaches \cite{vestner2017efficient, lai2017multiscale, melzi2019zoomout} just adopt an empirical or prefixed  increasing sequence.  

Motivated from the above difficulties and limitations of iterative refinement approaches in spectral domain, we propose a dual iterative refinement method which iteratively updates on the spatial and spectral domains. The first key idea is to choose well-matched pairs using local spatial matching from current correspondence, called anchor pairs, and only use them to determine the functional map which helps to construct a much more improved correspondence for the next iteration. In order to choose high-quality anchor pairs from a given correspondence, we zoom in at each corresponding pair and measure local mapping distortion. This step integrates local spatial information in the spectral refinement process. The second key idea is to find the maximal dimension of spectral embedding that can be robustly and accurately determined by these anchor pairs according to their distribution in spectral domain using singular value analysis on their correlation matrix.

\subsection{Local mapping distortion}
\label{sec:localdistortion}
In order to choose well-matched anchor pairs, or to filter out bad correspondences, to compute the functional map $C$ more accurately, we use the following \textit{local mapping distortion} (LMD) introduced in \cite{xiang2020efficient} to measure the matching quality of a given correspondence pair and choose those pairs with small LMD. Without relying on any information of the ground truth correspondence, this distortion provides a quantitative measurement to check the accuracy of the map through its continuity and local distance preservation,

\begin{definition}[Local mapping distortion (LMD)]
\label{def:faithfulness}
Let $T:\M_1 \rightarrow \M_2$ be a map between two manifolds. For any point $\x\in\M_1$, consider its $\gamma$-geodesic ball in $\M_1$ as $\B_\gamma(\x) = \{\y\in\M_1~|~ d_{\M_1}(\x,\y)\leq \gamma\}$. LMD of $T$ at $\x$ is defined as:
\begin{equation}
    \D_{\gamma}(T)(\x) = \frac{1}{|\B_\gamma(\x)|} \int_{\y\in \B_\gamma(\x)} DE_T(\x,\y)\mathrm{d} \y 
\label{eq:faithfulness}
\end{equation}
where  
$\displaystyle DE_T(\x,\y) = |d_{\M_1}(\x,\y)-d_{\M_2}(T(\x),T(\y))|/\gamma $ 
 and $|\B_\gamma|$ is the volume of $\B_\gamma$. 
\end{definition}

Based on the above definition of LMD, it is straightforward to check if
 $T$ is an isometric map, then $\D_{\gamma}(T)(\x) = 0, \forall \x\in\M_1, \gamma>0$.
Conversely, if $\D_{\gamma}(T)(\x) = 0, \forall \x\in\M_1$ for some $\gamma>0$, then $T$ is isometric.

In discrete setting, equation \eqref{eq:faithfulness} can be approximated by:
\begin{equation}
     \D_{\gamma}(T)(\x_i) \approx \frac{\sum_{\x_j \in \B_\gamma(\x_i)} \A_1(j) DE_T(\x_i,\x_j)}{  \left(\sum_{\x_j \in \B_\gamma(\x_i) } \A_1(j) \right)}
    \label{eq:test} 
\end{equation}
where 
$\A_1$ is the area element of the mesh representing $\M_1$.  

\begin{figure}[h]
    \centering
    \includegraphics[width = .45\linewidth]{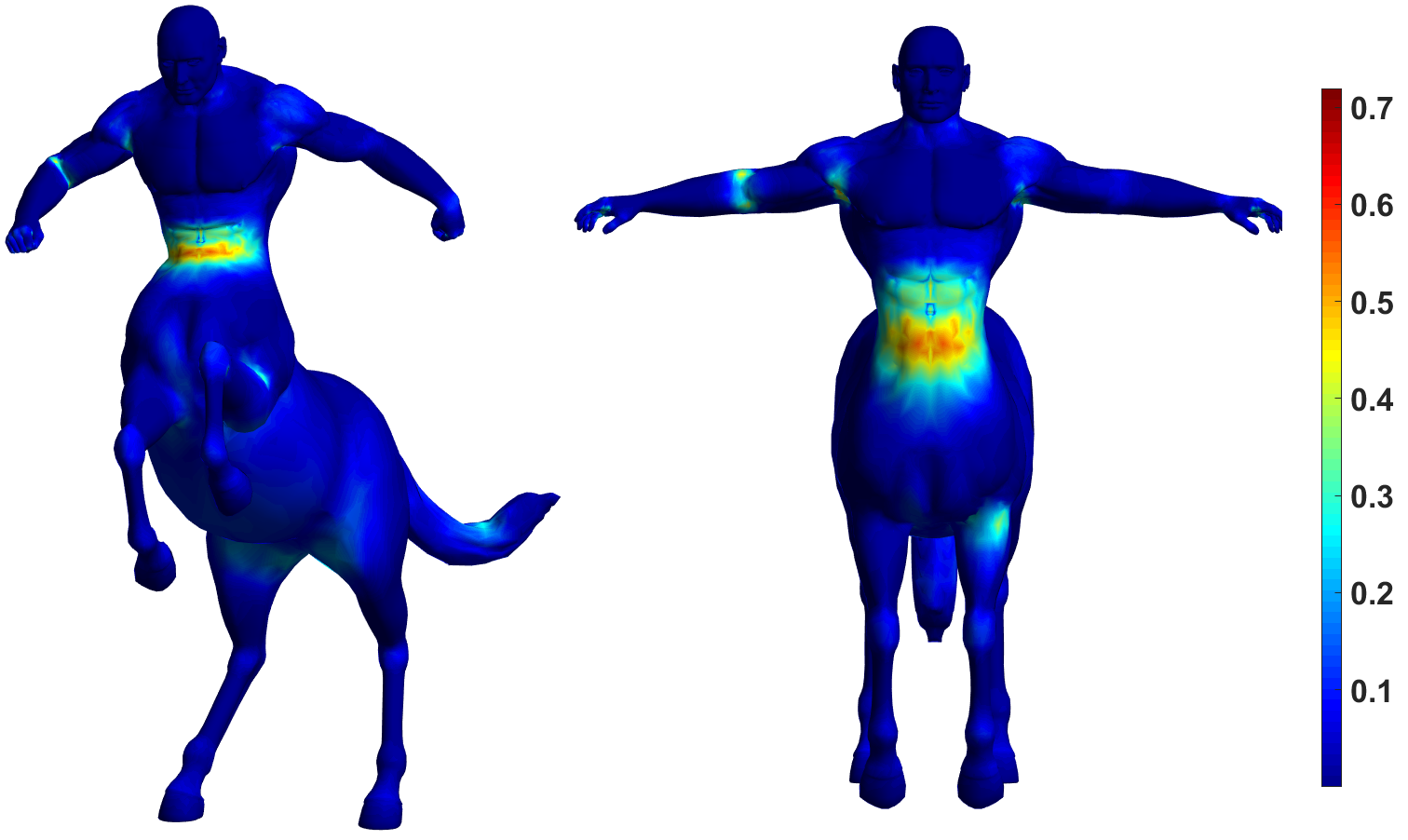}
    \caption{LMD on ground truth correspondence.}
    \label{ground_ld}
\end{figure}
By normalizing the distance with respect to $\gamma$, LMD is robust to global scaling as well as local sampling variation. Figure \ref{ground_ld} shows that, for the ground truth correspondence, LMD is very small almost everywhere except in regions where non-isometric distortion is large. Hence the proposed LMD also serves as a good unsupervised error metric when ground truth is not available.

\subsection{Our Method}
\label{subsec:ourmethod}
We first precompute $K$ leading LB eigen-functions for each shape. Here $K$ can be determined by computation cost limitation, the noise level in the data, or desired accuracy of the correspondence. In any case, it should not exceed what the mesh resolution can support for the discretized shape, i.e., a few mesh points are needed in each nodal domain. 

Our method can start with an initial correspondence provided by SHOT \cite{tombari2010unique} based on extrinsic point-wise features, or a few given landmarks used as initial anchor pairs which can be fixed or updated in later refinement, or by any other (fast but not necessary very accurate) methods. Then we start DIR which includes the following three simple steps: 
\textcircled{1} Choose anchor pairs from current correspondence using LMD criterion;
\textcircled{2} Determine the spectral dimension and the corresponding functional map based on the selected anchor pairs;
\textcircled{3} Update the correspondence using the current functional map.
In addition, we enforce two stopping criteria, the total number of iterations and the spectral dimension supported by anchor pairs reaches $K$,  whichever is satisfied first. 
Here are more detailed descriptions of each step.

\noindent\textbf{Step 1: Selecting anchor pairs.}

By setting a proper LMD threshold $\epsilon$, the set of pairs $\{(x_\ell, T(x_\ell)) | \D_{\gamma}(T)(\x_\ell) < \epsilon\}$ are selected as anchor pairs from the current correspondence which will be used to guide next refinement. It is important to note that both the threshold $\epsilon$ and anchor pairs are updated in later refinement. By decreasing $\epsilon$ with iterations, the quality of anchor pairs is also improved. 

\noindent\textbf{Step 2:} There are two components described as follows.

\noindent(1) \textbf{Determining the proper spectral dimension based on anchor pairs.}
Given a set of anchor pairs, an important question is to find a proper spectral dimension that can be determined accurately and stably according to the distribution of the anchor pairs in spectral embedding. Due to possible degeneracy of the distribution, the dimension can be quite smaller than the number of anchor pairs.
We use singular value decomposition (SVD)~\cite{trefethen1997numerical}, a.k.a principal component analysis, to find the dimension well expanded by a set of anchor pairs in spectral domain. 
\begin{equation}
   U \Sigma V^\top =  \Phi_{\M_1}^{K}(\{ x_\ell\}_{\ell=1}^m)^\top \Phi_{\M_2}^{K}(\{T( x_\ell)\}_{\ell=1}^m). 
\end{equation}
We threshold the singular values to determine the proper dimension. One can use a simple thresholding strategy. In our implementation, we adopt a normalization strategy to make the threshold adaptive to the data and noise level. After normalizing all singular values by the mean of 10 largest singular values, the dimension cut is set at where the sum of ten consecutive normalized singular values is smaller than a threshold (0.1 for all our tests). We point out that since the spectral embedding is defined by each eigen-function of the LB operator, instead of the collectively spanned space, the accuracy and stability of each singular vectors is more relevant in our case. For example, perturbation analysis for SVD can be found in \cite{liu2008first}.

\noindent(2) \textbf{Computing functional map based on anchor pairs.}
Once the proper spectral dimension, $k$, is determined by the selected anchor pairs $\{(x_\ell, T(x_\ell)~|~ \ell = 1,\cdots,m\}$, we compute the functional map for $k$ dimensional spectral embedding only based on the anchor pairs as follows:
\begin{equation}
\begin{split}
    C &= \arg\min_{C\in\fR} \sum_{\ell=1}^m \|\ \Phi^k_{\M_1}(x_\ell)C - \Phi^k_{\M_2}(T(x_\ell)) \|_F^2 \\
     &= \proj\left(\left(\Phi_{\M_1}^k(\{x_\ell\}_{\ell}) \right)^\top \Phi_{\M_2}^k(\{T (x_\ell)\}_{\ell})\right)
\end{split}
\end{equation}
This restricted version to compute a functional map minimizes possible corruption from inaccurate correspondence and leads to a more accurate estimation of the functional map as discussed in the Section \ref{subsec:FM} and  supplementary material. Meanwhile, it also reduces the computation cost. 

\noindent\textbf{Step 3: Construct the new correspondence.} Using the functional map computed from selected anchor pairs in the properly enlarged spectral embedding space, a refined correspondence is constructed by solving the assignment problem \eqref{eqn:knnmap}, where a KNN search method is applied. 



We summarize the full procedure in Algorithm~\ref{alg7}. 
\begin{algorithm}
\label{alg7} 
\caption{Dual Iterative Refinement (DIR)}
\textbf{input} An initial correspondence, $\Phi^K_{\M_1}$ and $\Phi^K_{\M_2}$ (e.g. $K$ leading LB eigenfunctions), LMD error  $\{\epsilon_i\}_{i=1}^N$ thresholds and the maximum iteration $N$.

\While{$i\leq N$, or $k_i \leq K$}{
 1. Find anchor pairs $\{  (x_\ell, T(x_\ell)) ~|~\D_{\gamma}(T)(x_\ell) < \epsilon_i \}_{\ell=1}^{ m_i}$ via LMD criterion from current correspondence $T$.

 2. Find spectral dimension $k_i$ from SVD of $\Phi_{\M_1}^{K}(\{ x_\ell\}_{\ell})^\top \Phi_{\M_2}^{K}(T(\{ x_\ell\}_{\ell}))$.

 3. Update the functional map $C = \proj\left(\left(\Phi_{\M_1}^{k_i}(\{x_\ell\}_{\ell}) \right)^\top \Phi_{\M_2}^{k_i}(\{T (x_\ell)\}_{\ell})\right)$

 4. Update the correspondence (for all $p\in\M_1$) $\displaystyle T(p) = \arg \min_{q} \| \Phi_{\M_1}^{k_i}(p,:) C - \Phi_{\M_2}^{k_i}(q,:) \|^2$.
} 
\end{algorithm}
\noindent\textbf{Computation complexity} The complexity for computing $K$ leading eigen-vectors of a $n\times n$ sparse matrix corresponding to the discretized LB operator is $O(Kn\log n)$. The complexity for checking LMD is $O(n)$ since geodesic distance is only computed in a fixed neighborhood, e.g., first ring or second ring, at each point.
To compute the functional map, the complexity of matrix multiplication is at most $O(nK^2)$ and SVD decomposition is at most $O(K^3)$ since the number of anchor pairs is at most $n$ and the spectral dimension $k$ is at most $K$ which is prefixed and far less than $n$. KNN search used to solve Equation~\eqref{eqn:knnmap} has a complexity of $O(n\log(n))$~\cite{vaidya1989ano}. Hence, our method is of complexity $O(n\log(n))$ altogether. It is still $O(n\log(n))$ using other global features such as geodesic distance to anchor points since we limit the maximal number of anchor points (as the number of LB eigen-functions). The complexity of computing geodesic distance and solving the resulting assignment problem is still $O(n\log(n))$.

\section{Discussion}
In this section, we discuss other possible extensions of our approach and a few specific applications. 

\noindent\textbf{Combination of local and global features.}
For shape correspondence problem, an efficient and robust approach should use both local and global features. 
So far we have mainly talked about using spatial and spectral features, which are perfectly complementary in the sense of local and global information.
Anchor pair selection by the LMD criterion is based on local spatial features while the spectral features are global. We would like to point out that the proposed strategy of DIR process can be extended to other combinations of appropriate local and global features in different applications. For examples, one may alternatively use Heat Kernel Signature \cite{bronstein2010scale}, Wave Kernel Signature \cite{aubry2011wave} or Geodesic Distance Signature (GDS) \cite{van2011survey} as global features. For shapes with holes and boundaries, we choose GDS which is less sensitive to local mesh distortions and boundaries for most interior points. However, if accurate spectral information is available, taking advantage of the multi-scale representation in spectral embedding leads to better accuracy and efficiency according to our tests.

\noindent\textbf{DIR with limited initial landmarks.}
In our previous discussion, the first collection of anchor pairs is selected from a given initial correspondence such as the one obtained from comparing SHOT features. Our method does enjoy the flexibility to incorporate given landmarks. This is applauded in applications like shape matching in the morphological study in medical imaging where landmarks could be annotated based on specific tasks~\cite{gu2004genus}. Once a few human-annotated and required landmarks are given, which are taken as the initial anchor pairs and fixed in later iterations, DIR will converge to a stable solution. The fewer landmarks are provided, the more iterations are usually needed. The final convergence performance of our numerical tests, as shown in Figure \ref{humanld}, indicates that DIR can provide accurate correspondence based on only four landmarks and it is stable with respect to different initialization.

\noindent\textbf{Matching of point clouds without mesh.}
In real applications, point clouds with well-constructed global triangulation is usually hard to obtain. For point clouds without global mesh, we use the local mesh method to compute the LB eigenvalues and LMD~\cite{lai2013local,xiang2020efficient}. DIR works well on matching raw point clouds as shown in section~\ref{sec:Experiments}. 

\noindent\textbf{Patch/partial matching.}
Patch/partial matching often comes with difficulties due to artificial boundaries, different sizes and topological perturbations. Spectral information is either unavailable or unreliable to use. However, our local feature, LMD, is not affected by the above difficulties at interior points. By replacing spectral features with geodesic distance to anchor points, which is global and less sensitive to the above difficulties for most interior points, DIR can handle patch or partial matching well as shown in our numerical experiments.



\section{Experiments}
\label{sec:Experiments}
In this section, we conduct comprehensive experiments to evaluate the performance of our method on various benchmark data sets.
In all experiments, no pre-processing, such as a low resolution model or pre-computing the geodesic distance matrix, is required in our algorithm. Raw point clouds (with or without meshes) are directly used. In our comparisons, all geodesic errors from existing methods are obtained from the associated error curve data appeared in the papers. Computation using ZoomOut is produced from the code on GitHub shared by the authors~\cite{melzi2019zoomout}(3 to 150 spectral basis). All experiments using our methods are conducted in Matlab with 16GB RAM and Intel i7-6800k CPU.

\noindent\textbf{Error Metric.}~~We use the geodesic error as our error metric in most experiments. Suppose the constructed correspondence maps $\x\in \M_1$ to $\y \in \M_2$ while the true correspondence is $\x$ is to $\y^*$, we measure the quality of our result by computing the geodesic error defined as $e(x) = \frac{d_{\M_2}(y,y^*)}{diam(\M_2)}$, where $diam(\M_2)$ is the geodesic diameter of $\M_2$.

\noindent\textbf{Hyperparameters.}~~We use the second ring  neighborhood size for LMD criterion; maximum iteration number is 10 with the LMD threshold as $[0.26, 0.22, 0.18, 0.14, 0.1, 0.1, 0.1, 0.1, 0.1, 0.1]$. The initial correspondence is given by the KNN search result based on SHOT features \cite{tombari2010unique}. These hyperparameters, especially the LMD, are not sensitive to different data sets. We use the same hyperparameters on TOSCA and SCAPE data sets. For SHREC'16 and patch matching, we set the maximum 800 anchor pairs served as the reference points for geodesic distance features. 

\noindent\textbf{TOSCA~\cite{bronstein2008numerical}, SCAPE~\cite{anguelov2005correlated}.}~TOSCA data set consists of 76 shapes in 8 different categories ( human and animal shapes) with vertex numbers ranging from 4k to 50k. We present the results of our method using mesh structure with the maximal spectral dimension as 1000 and 500 respectively. We also apply our method using only point clouds (no mesh) with the maximal spectral dimension as 1000 (using local mesh method~\cite{lai2013local}). We perform the same experiments on SCAPE data set which has 72 shapes (12,500 vertices for each) of the same person with different poses. 

\begin{figure}[h]
\centering\includegraphics[width=.75\linewidth]{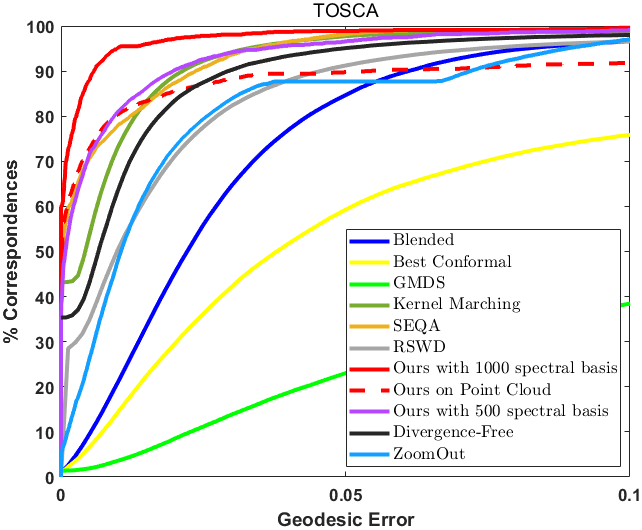}
\centering\includegraphics[width=.75\linewidth]{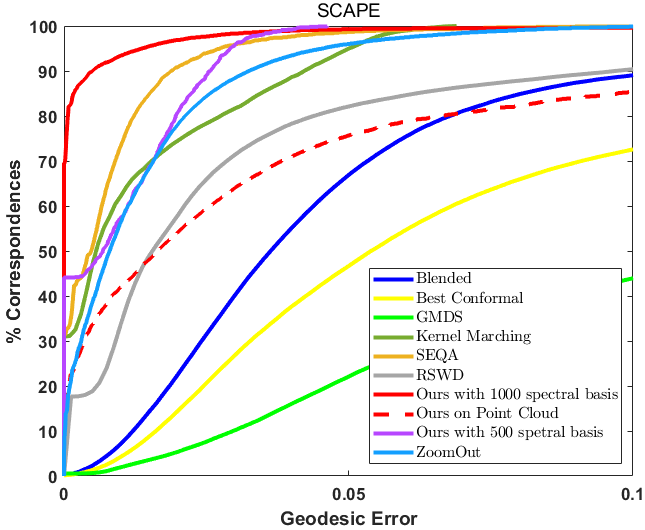}

\caption{Geodesic error on the TOSCA and SCAPE data sets with comparison to other methods.}
\label{TSres}
\end{figure}

In Figure~\ref{TSres}, We compare our method with the following methods: Blended \cite{kim2011blended}, Best Conformal \cite{kim2011blended}, GMDS \cite{bronstein2006generalized}, Kernel Marching \cite{vestner2017efficient}, SEQA \cite{xiang2020efficient}, RSWD \cite{lai2017multiscale}, ZoomOut \cite{melzi2019zoomout}, Divergence-Free Shape Interpolation \cite{eisenberger2018divergence}, SRFM~\cite{ren2019structured} and BCICP~\cite{ren2018continuous}. Our method with 1000 spectral basis outperforms all methods, our method with 500 spectral basis outperforms almost all methods, and our method on purely point cloud inputs still achieves good performance. 

Table~\ref{runtime} indicates the computation efficiency of our methods by showing the average run time on several examples from TOSCA including shapes with vertices ranging from 4k to 50k. CUP time of our method reports total time of all steps including computation of LB eigenfunctions.  As discussed in Section~\ref{subsec:ourmethod}, the complexity of our method is $O(n\log(n))$. But in practice, when computing shapes with more than 20,000 vertices, our computer suffers a computation speed slow-down from the vast RAM usage because of our limited RAM capacity. Hence, for Cat and David shape from TOSCA, the run time is higher than expected. Most state-of-the-art approaches have a computation complexity of $O(n^2)$ and do not report run time for shapes over 10,000 vertices. Our method is very efficient compared with state-of-the-art methods which report computation time. 
\begin{table}[h]
\centering
\scalebox{0.7}{
\begin{tabular}{c|c|c|c|c|c}
\hline
Model                                                                              & Wolf & Centaur & Horse & Cat   & David \\ \hline
Number of Vertices                                                                 & 4344 & 15768   & 19248 & 27894 & 52565 \\ \hline
SEQA (s)                                                                           & 59   & 531     & 801   & 929   & 1681  \\
Kernel Matching (s)                                                                & 60   & NA      & NA    & NA    & NA    \\
ZoomOut(3-150 spectral basis) (s)                                            & 25   & 277      & 449    & 1267    & 4312    \\
Consistent Shape Matching (s)                                                      & 483  & NA      & 2118  & 3299  & NA    \\ \hline
\begin{tabular}[c]{@{}c@{}}Our Method with \\ 1000 spectral basis (s)\end{tabular} & 36   & 144     & 191   & 474   & 1774  \\
\begin{tabular}[c]{@{}c@{}}Our Method with \\ 500 spectral basis (s)\end{tabular}  & 19   & 106     & 131   & 330   & 1301  \\ \hline
\end{tabular}}
\vspace{0.2cm}
\caption{Average run time for shapes from TOSCA comparing with SEQA~\cite{xiang2020efficient}, ZoomOut~\cite{melzi2019zoomout}, Consistent Shape Matching~\cite{azencot2019consistent} (without including precomputation time for the geodesic distance matrix) and Kernel Matching~\cite{vestner2017efficient}.}
\label{runtime}
\vspace{-0.3cm}
\end{table}

\noindent\textbf{SHREC'16~\cite{cosmo2016matching}}~~ SHREC'16 Partial Correspondence benchmark data set consists of 8 types of isometric human or animal shapes in different poses with regular ‘cuts’ and irregular ‘holes’. We test our method by matching each partial shape to the corresponding full shape. Since spectral basis is sensitive to mesh 'cuts' and 'holes', we use geodesic distance to anchor points as global features. We still use SHOT for the initialization.
In Figure~\ref{SHREC'16}, We compare our method with ZoomOut \cite{melzi2019zoomout}, Partial Functional Maps \cite{rodola2017partial} and Random Forests \cite{rodola2014dense}. 
The results show that our method is quite flexible and robust to handle shape matching for different scenarios. It again outperforms other state-of-the-art methods markedly and achieves high accuracy on this challenging data set.

\begin{figure}[h]
\vspace{-0.3cm}
    \centering
    \includegraphics[width = .9\linewidth]{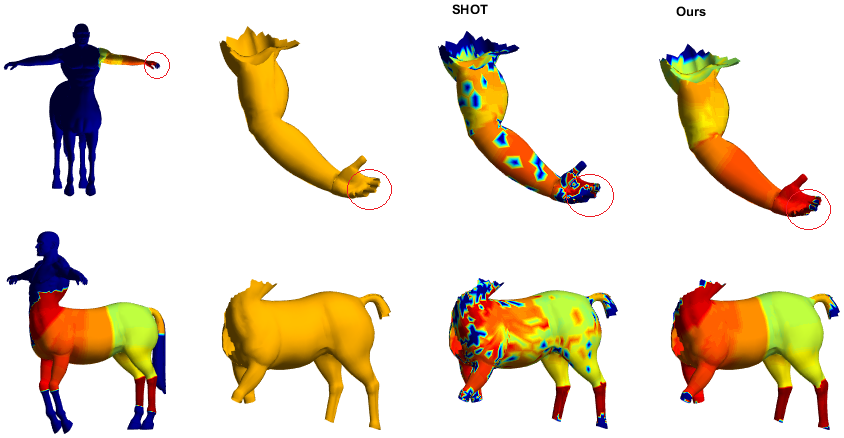}
    \caption{Partial matching between body parts in the second column and the entire body. The first column: the entire centaurs with the ground truth color map (non-blue area). Extra points are colored in blue. Removed finger tips are highlighted by red circles. The third and fourth columns: the color map of a result using SHOT and our method, respectively.}
    \label{patch}
\end{figure}

\noindent\textbf{Patch Matching.}~~ Since there is no standard data set for patch matching, so instead we report several experiments using patches cut from TOSCA data set. The first case contains two examples with artificial boundaries, different sizes (partial patching) and topological changes simultaneously. In the first example, we take a portion of an arm (finger tips also removed) from one of two nearly isometric centaur shapes, and then map the partial arm onto the other entire centaur. In the second example, we map a body patch onto the whole shape. The original centaur is a closed mesh surface with no holes, while the arm and the body patch are not closed and have holes and boundaries. Our method performs quite well even for this challenging example as shown in Figure \ref{patch}. 

\begin{figure}[h]
\vspace{-0.6cm}
    \centering
    \includegraphics[width = .9\linewidth]{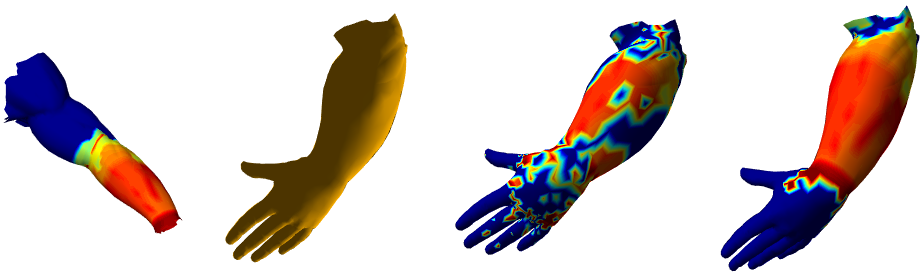}
    \setlength{\belowcaptionskip}{-8pt} 
    \caption{Patch matching. From left to right: the source patch (with the ground truth color map in the non-blue area and extra points colored in blue), the target patch, mapping result using SHOT and mapping result using our method. Blue colored regions in columns three and four indicate points not passing the post LMD test.}
    \label{overlap}
\end{figure}

The second case is matching two patches containing both overlap and non-overlap parts. We match an arm without the hand to a portion of an arm with the hand, where forearm is the common part. Since there is no correspondence between the non-overlap parts, a post LMD test is added to prune out those points. The result is shown in Figure \ref{overlap}. Again our method performs quite well. These experiments indicate that our method is robust to size differences, artificial boundaries and topological changes.

\begin{figure}
\vspace{-0.3cm}
    \centering
\centering\includegraphics[width=.4\linewidth]{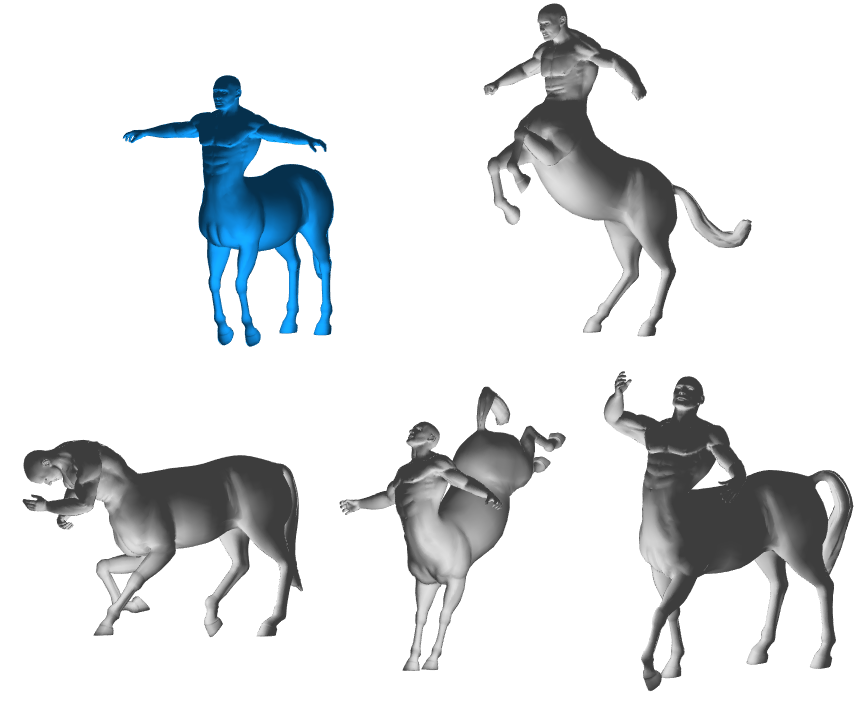}
\centering\includegraphics[width=.51\linewidth]{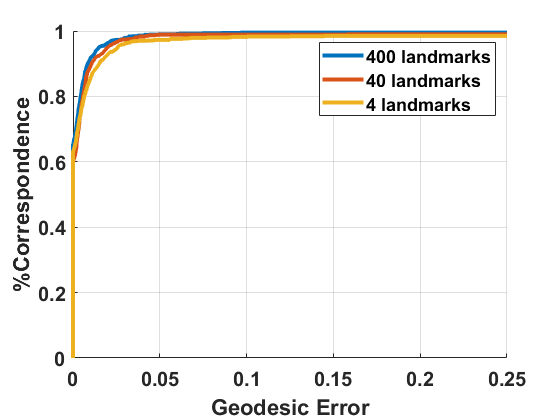}
    \caption{Average performance of our method on several pairs of centaur shapes from TOSCA (blue centaur map to gray centaurs), given different number of initial random annotated landmarks.}
    \label{humanld}
    \vspace{-0.2cm}
\end{figure}

\noindent\textbf{Experiments Using Limited Landmarks.}~~
We select 4 pairs of centaur shapes from TOSCA, initialize our method with different number of randomly chosen landmark pairs (without using SHOT for initialization) and then plot the average geodesic error curve. These landmark pairs are fixed and always belong to our selected anchors pairs in later iterations. After different number of iterations, the final performance is illustrated in Figure~\ref{humanld}. Even with four initial random landmarks, the performance is excellent although more iterations are needed. It shows stability and robustness of our method with respect to initial correspondence. 

\begin{figure}[h]
\centering
\begin{minipage}{0.4\linewidth}
\includegraphics[width=1\linewidth]{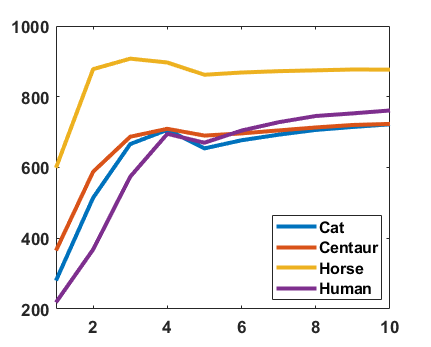}
\end{minipage}
\begin{minipage}{0.59\linewidth}
\includegraphics[width=1\linewidth]{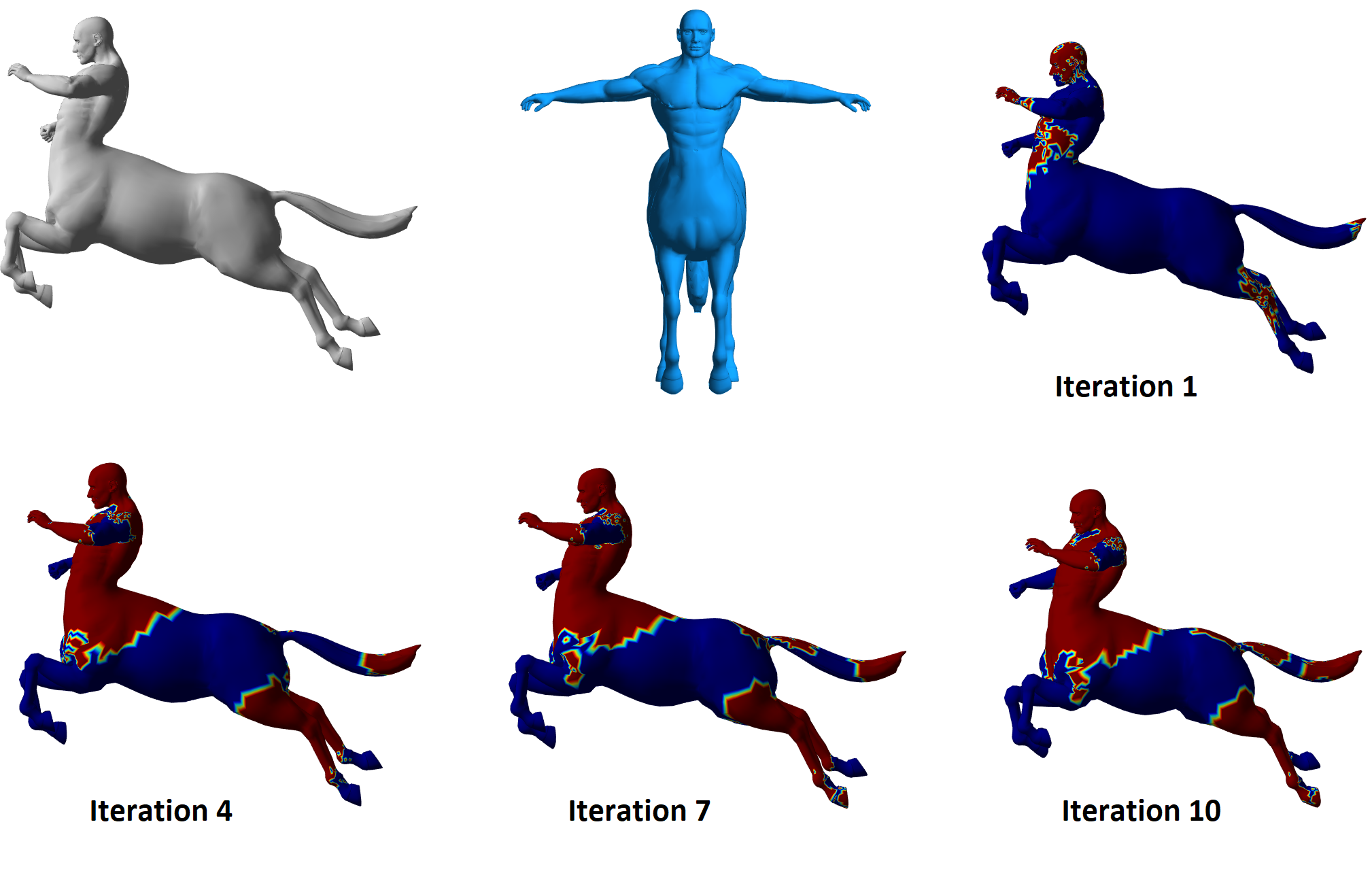}
\end{minipage}\\
\vspace{0.2cm}
\caption{Left plot: The average spectral dimension (with a maximal number of LB eigenfunctions set to 1000) for cat, horse, centaur class from TOSCA and human from SCAPE at each iteration. Righ figurest: Selected anchor pairs (color-coded in red) during iterations of finding the correspondence between the gray centaur and the blue one. The anchor pairs detected at iteration 1 are directly from matching SHOT descriptors.}
\label{noe}
\vspace{-0.3cm}
\end{figure}

\noindent\textbf{Growth of Anchor Pairs and Spectral Dimensions.}~~ One key feature of our method is the selection of well matched pairs used to determine spectral dimension and the corresponding functional map at next iteration in an automatic and data-adaptive way. A steady growth and improvement (by decreasing the mapping distortion threshold with iterations) of anchor pairs (updated in each iteration) means a steady growth of the spectral dimension and increase of resolution which leads to an effective and fast refinement of the correspondence. 
The left image in Figure~\ref{noe} shows the average spectral dimensions for several class of shapes from TOSCA or SPACE data set. The right image in Figure~\ref{noe} shows the growth pattern of anchor pairs. These indicate that our method can extract anchor pairs and increase spectral resolution in a steady and efficient way.

\section{Conclusion}
\label{sec:conclusion}
We propose a simple and efficient iterative refinement strategy utilizing dual features, spatial and spectral or local and global. By only relying on selected well-matched pairs to guide the next refinement, our proposed method combines complementary information in an optimal and data-adaptive way. Our method shows superior performance on extensive tests compared to other state-of-the-arts methods in terms of accuracy, efficiency and stability.

\section*{Acknowledgement}
R. Lai is supported in part by an NSF Career Award DMS--1752934, and H. Zhao is supported by an NSF Award DMS-1821010.


{\small
\bibliographystyle{ieee_fullname}
\bibliography{DIRbib}
}

\clearpage
\appendix

\section{The Effect of Correspondence Error on Spectral Alignment}
\label{proof_of_th1}

\textbf{Proof of Theorem \ref{thm:Error}} We first show that a necessary condition for $\|C_a - C_T\|_2 = 0$ is that permutation $\sigma$ is an involution, i.e. symmetric matrix. 

Since $\Phi_{\M_2} = \Phi_{\M_1} C_T$, we have $Y_i = X_i C, i = 1, 2$. It is clear that $\Phi_{\M_1}^\top \Pi \Phi_{\M_2} = (X_1^\top X_1 + X_2^\top \sigma X_2)C_T$. Assume that there is a SVD decomposition $X_1^\top X_1 + X_2^\top \sigma X_2 =  U\Sigma V^\top$. Then we have a SVD decomposition $\Phi_{\M_1}^\top \Pi \Phi_{\M_2} = U\Sigma V^\top C_T$ because $C_T$ is an orthonormal matrix. Therefore, $C_a = \proj(\Phi_{\M_1}^\top \Pi \Phi_{\M_2}) = UV^\top C_T$. This yields 
\begin{equation*}
    \|C_a - C_T\|_2 = \|UV^\top C_T - C_T\|_2 = \|UV^\top - \mathrm{Id}\|_2
\end{equation*}
Thus, $\|C_a - C_T\|_2 = 0$ implies $U = V$. This leads to $\sigma$ is symmetric. From the fact that there are $n_2!$ permutation matrix of size $n_2\times n_2$ and there are $\displaystyle n_2!\sum_{j=0}^{\lfloor n_2/2 \rfloor}\frac{1}{2^j j!(n_2-2j)!}$ symmetric permutation matrix of size $n_2\times n_2$~\cite{skiena1990}. This concludes the proof.

\vspace{0.5cm}
One obvious observation is that $\eta=\displaystyle \sum_{j=0}^{\lfloor n_2/2 \rfloor}\frac{1}{2^j j!(n_2-2j)!}$ decreases rapidly as $n_2$ grows. For example, when $n_2 = 25$, $\eta \approx 10^{-12}$. However, to give a more quantitative characterization of the perturbation for an arbitrary shuffling is difficult since it depends not only on $\sigma$ but also on $X_1$ and $X_2$.
Instead, we conduct a few numerical experiments to demonstrate how an inaccurate correspondence $\Pi$ will affect the correlation matrix $\Phi_{\M_1}^\top \Pi \Phi_{\M_2}$, which is the essential information to align spectral basis, e.g., functional map, between two shapes. 

\begin{figure}
    \centering
    \includegraphics[width = .9\linewidth]{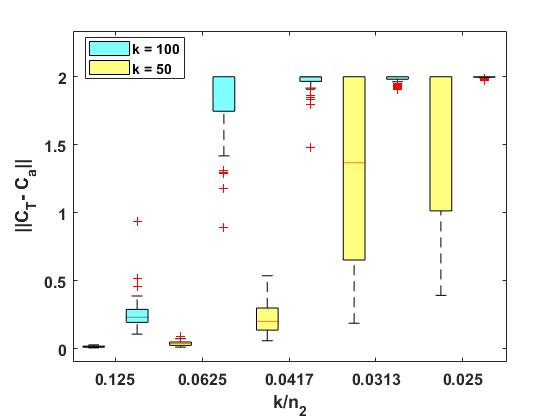}\\
    \centering
    \includegraphics[width = .9\linewidth]{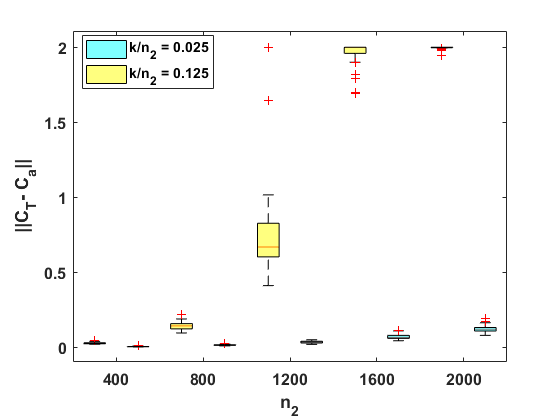}
    \caption{\textit{Top:} Error statistics for the relation between functional map error and $\frac{k}{n_2}$. \textit{Bottom:} Error statistics for the relation between functional map error and number of points in $X_2$.}
    \label{th1_exp}
\end{figure}

In our experiments, $\Phi_{\M_1}, \Phi_{\M_2} \in \mathbb{R}^{n\times k}$, where $n$ is the total number of points and $k$ is the spectral dimension. 
We map a human shape with 12,500 vertices to itself, which is a perfect isometric shape correspondence problem with identity map as the ground truth. Theoretically, $0 \leq \|C_a - C_T\|_2 \leq 2$, since both $C_a$ and $C_T$ are orthonormal. 
The first experiment tests the behavior of $|| \C_T - \C_a||_2$ with respect to the ratio of $\frac{k}{n_2}$ for two fixed spectral dimension, $k=50$ and $k=100$; second experiments tests the behavior of $|| \C_T - \C_a||_2$ with respect to $n_2$ for two fixed $\frac{k}{n_2}$ ratio. A random permutation is imposed on $Y_2$ to compute $C_a$, and we independently run 50 trails for each parameter combination. Box-plots are used to illustrate the statistics of our computation in Figure \ref{th1_exp}.
The experiments indicate that using inaccurate correspondences at the current step will introduce error to the estimation of functional map at the next step and the error can be quite significant (as big as the worst case, $\|C_a - C_T\|_2 = 2$). This could cause a failure or slow convergence for an iterative refinement strategy. 
 

\begin{figure}
    \centering
    \includegraphics[width = .25\linewidth]{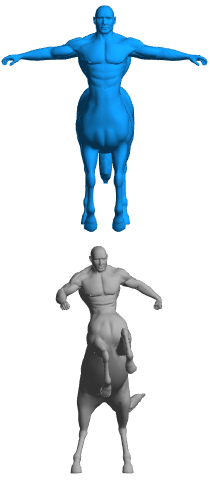}
    \includegraphics[width = .74\linewidth]{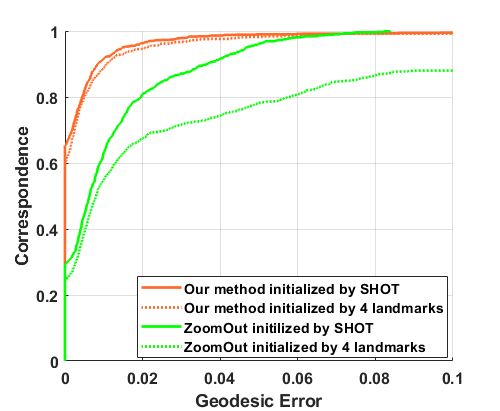}
    \caption{\textit{Left: } The blue centaur is mapped to the gray centaur from TOSCA data set. \textit{Right: } Geodesic error of our method and ZoomOut using different initial guess.} 
    \label{bad_exp}
\end{figure}

We further test a real example on one pair of shapes from TOSCA data set using ZoomOut~\cite{melzi2019zoomout}, which is a simple iterative refinement method based on functional map in spectral domain. A fixed increment of spectral dimension is pre-specified for each iteration. Due to the use of current correspondence at all points to construct the functional map, the following iterative refinements may not lead to a satisfactory result in the end. In this test, we use 1000 LB eigen-functions for ZoomOut (and our method) with two different initial setups: (1) correspondence provided by SHOT, or (2) 4 given landmarks. For ZoomOut, we first compute a functional map for the first 4 spectral modes from the initial correspondence and then use the code provided by the authors on GitHub to run the experiment. Test results are plotted in Figure~\ref{bad_exp}.
These experiment further verify our analysis on possible issues using simple iterative spectral refinement without filtering out bad correspondences. 

\end{document}